\documentclass[11pt,a4paper]{article}
\usepackage[hyperref]{emnlp2020}
\usepackage{times}
\usepackage{latexsym}
\usepackage{xcolor}
\usepackage{booktabs}
\usepackage{amsmath}
\usepackage{amssymb}
\usepackage{bbm}
\usepackage{enumitem}
\usepackage{algpseudocode}
\usepackage[noend,ruled]{algorithm2e}
\usepackage{array,multirow,graphicx}
\usepackage{float}
\usepackage{subfigure}
\usepackage{longtable}
\usepackage{supertabular}
\newcommand{\bs}[1]{\boldsymbol{#1}}
\newcommand{\red}[1]{\textcolor{red}{#1}}

\newcommand{\ie}{\emph{i.e.}}
\newcommand{\eg}{\emph{e.g.}}
\newcommand{\lotclass}{\textbf{LOTClass}\xspace}
\definecolor{midnightgreen}{rgb}{0.0, 0.29, 0.33}

\definecolor{RoseQuartzBg}{HTML}{F7CAC9}
\definecolor{RoseQuartz}{HTML}{F5A798}
\definecolor{Serenity}{HTML}{92A8D1}
\definecolor{OrangeRed}{rgb}{1.0, 0.27, 0.0}
\definecolor{Turquoise}{HTML}{0F4C81}
\usepackage{xparse}
\NewDocumentCommand{\heng}{ mO{} }{\textcolor{OrangeRed}{\textsuperscript{\textit{Heng}}\textsf{\textbf{\small[#1]}}}}

\NewDocumentCommand{\yu}{ mO{} }{\textcolor{blue}{\textsuperscript{\textit{Yu}}\textsf{\textbf{\small[#1]}}}}

\def\aclpaperid{1611}

\usepackage{microtype}
\usepackage{makecell}

\aclfinalcopy 

\title{
Text Classification Using Label Names Only: A Language Model Self-Training Approach
}

\author{Yu Meng$^{1}$, Yunyi Zhang$^{1}$, Jiaxin Huang$^{1}$, Chenyan Xiong$^{2}$, \\ \textbf{Heng Ji$^{1}$, Chao Zhang$^{3}$, Jiawei Han$^{1}$} \\
  $^1$University of Illinois at Urbana-Champaign, IL, USA \\
  $^2$Microsoft Research, WA, USA  \ \ \ $^3$Georgia Institute of Technology, GA, USA \\
  $^1$\texttt{\{yumeng5, yzhan238, jiaxinh3, hengji, hanj\}@illinois.edu} \\ $^2$\texttt{chenyan.xiong@microsoft.com} \ \ \ $^3$\texttt{chaozhang@gatech.edu} \\
  }

\date{}

\begin{document}
\maketitle

\begin{abstract}


Current text classification methods typically require a good number of human-labeled documents as training data, which can be costly and difficult to obtain in real applications. Humans can perform classification without seeing any labeled examples but only based on a small set of words describing the categories to be classified. In this paper, we explore the potential of only using the label name of each class to train classification models on unlabeled data, without using any labeled documents. We use pre-trained neural language models both as general linguistic knowledge sources for category understanding and as representation learning models for document classification. Our method (1) associates semantically related words with the label names, (2) finds category-indicative words and trains the model to predict their implied categories, and (3) generalizes the model via self-training. We show that our model achieves around $90\%$ accuracy on four benchmark datasets including topic and sentiment classification without using any labeled documents but learning from unlabeled data supervised by at most $3$ words ($1$ in most cases) per class as the label name\footnote{Source code can be found at \url{https://github.com/yumeng5/LOTClass}.}. 

\end{abstract}


\section{Introduction}

Text classification is a classic and fundamental task in Natural Language Processing (NLP) with a wide spectrum of applications such as question answering~\cite{Rajpurkar2016SQuAD10}, spam detection~\cite{Jindal2007ReviewSD} and sentiment analysis~\cite{Pang2002ThumbsUS}.
Building an automatic text classification model has been viewed as a task of training machine learning models from human-labeled documents.
Indeed, many deep learning-based classifiers including CNNs~\cite{Kim2014ConvolutionalNN,Zhang2015CharacterlevelCN} and RNNs~\cite{Tang2015DocumentMW,Yang2016HierarchicalAN} have been developed and achieved great success when trained on large-scale labeled documents (usually over tens of thousands), thanks to their strong representation learning power that effectively captures the high-order, long-range semantic dependency in text sequences for accurate classification. 

Recently, increasing attention has been paid to semi-supervised text classification which requires a much smaller amount of labeled data. 
The success of semi-supervised methods stems from the usage of abundant unlabeled data: Unlabeled documents provide natural regularization for constraining the model predictions to be invariant to small changes in input~\cite{Chen2020MixTextLI,Miyato2017AdversarialTM,Xie2019UnsupervisedDA}, thus improving the generalization ability of the model.
Despite mitigating the annotation burden, semi-supervised methods still require manual efforts from domain experts, which might be difficult or expensive to obtain especially when the number of classes is large.

Contrary to existing supervised and semi-supervised models which learn from labeled documents, a human expert will just need to understand the label name (\ie, a single or a few representative words) of each class to classify documents. For example, we can easily classify news articles when given the label names such as ``sports'', ``business'', and ``politics'' because we are able to understand these topics based on prior knowledge.

In this paper, we study the problem of weakly-supervised text classification where only the label name of each class is provided to train a classifier on purely unlabeled data.
We propose a language model self-training approach wherein a pre-trained neural language model (LM)~\cite{Devlin2019BERTPO,Peters2018DeepCW,Radford2018ImprovingLU,Yang2019XLNetGA} is used as both the general knowledge source for category understanding and feature representation learning model for classification. The LM creates contextualized word-level category supervision from unlabeled data to train itself, and then generalizes to document-level classification via a self-training objective.

Specifically, we propose the \lotclass model for \textbf{L}abel-Name-\textbf{O}nly \textbf{T}ext \textbf{Class}ification built in three steps: (1) We construct a category vocabulary for each class that contains semantically correlated words with the label name using a pre-trained LM. (2) The LM collects high-quality category-indicative words in the unlabeled corpus to train itself to capture category distinctive information with a contextualized word-level category prediction task. (3) We generalize the LM via document-level self-training on abundant unlabeled data.

\lotclass achieves around $90\%$ accuracy on four benchmark text classification datasets, \textit{AG News}, \textit{DBPedia}, \textit{IMDB} and \textit{Amazon} corpora, \emph{without} learning from any labeled data but only using at most $3$ words ($1$ word in most cases) per class as the label name, outperforming existing weakly-supervised methods significantly and yielding even comparable performance to strong semi-supervised and supervised models. 

The contributions of this paper are as follows:
\begin{itemize}[leftmargin=*]
\item We propose a weakly-supervised text classification model \lotclass based on a pre-trained neural LM without any further dependencies\footnote{Other semi-supervised/weakly-supervised methods usually take advantage of distant supervision like Wikipedia dump~\cite{Chang2008ImportanceOS}, or augmentation systems like trained back translation models~\cite{Xie2019UnsupervisedDA}.}. \lotclass does not need any labeled documents but only the label name of each class.
\item We propose a method for finding category-indicative words and a contextualized word-level category prediction task that trains LM to predict the implied category of a word using its contexts. The LM so trained generalizes well to document-level classification upon self-training on unlabeled corpus.
\item On four benchmark datasets, \lotclass outperforms significantly weakly-supervised models and has comparable performance to strong semi-supervised and supervised models.
\end{itemize}


\section{Related Work}

\subsection{Neural Language Models}
Pre-training deep neural models for language modeling, including autoregressive LMs such as ELMo~\cite{Peters2018DeepCW}, GPT~\cite{Radford2018ImprovingLU} and XLNet~\cite{Yang2019XLNetGA} and autoencoding LMs such as BERT~\cite{Devlin2019BERTPO} and its variants~\cite{Lan2020ALBERTAL,Lewis2020BARTDS,Liu2019RoBERTaAR}, has brought astonishing performance improvement to a wide range of NLP tasks,
mainly for two reasons: (1) LMs are pre-trained on large-scale text corpora, which allow the models to learn generic linguistic features~\cite{Tenney2019WhatDY} and serve as knowledge bases~\cite{Petroni2019LanguageMA}; and (2) LMs enjoy strong feature representation learning power of capturing high-order, long-range dependency in texts thanks to the Transformer architecture~\cite{Vaswani2017AttentionIA}.

\subsection{Semi-Supervised and Zero-Shot Text Classification}

For semi-supervised text classification, two lines of framework are developed to leverage unlabeled data.
Augmentation-based methods generate new instances and regularize the model's predictions to be invariant to small changes in input. The augmented instances can be either created as real text sequences~\cite{Xie2019UnsupervisedDA} via back translation~\cite{Sennrich2016ImprovingNM} or in the hidden states of the model via perturbations~\cite{Miyato2017AdversarialTM} or interpolations~\cite{Chen2020MixTextLI}.
Graph-based methods~\cite{Tang2015PTEPT,Zhang2020MinimallySC} build text networks with words, documents and labels and propagate labeling information along the graph via embedding learning~\cite{Tang2015LINELI} or graph neural networks~\cite{Kipf2017SemiSupervisedCW}.

Zero-shot text classification generalizes the classifier trained on a known label set to an unknown one without using any new labeled documents. Transferring knowledge from seen classes to unseen ones typically relies on semantic attributes and descriptions of all classes~\cite{Liu2019ReconstructingCN,Pushp2017TrainOT,Xia2018ZeroshotUI}, correlations among classes~\cite{Rios2018FewShotAZ,Zhang2019IntegratingSK} or joint embeddings of classes and documents~\cite{Nam2016AllinTL}.
However, zero-shot learning still requires labeled data for the seen label set and cannot be applied to cases where no labeled documents for any class is available.

\subsection{Weakly-Supervised Text Classification}

Weakly-supervised text classification aims to categorize text documents based only on word-level descriptions of each category, eschewing the need of any labeled documents. Early attempts rely on distant supervision such as Wikipedia to interpret the label name semantics and derive document-concept relevance via explicit semantic analysis~\cite{Gabrilovich2007ComputingSR}. Since the classifier is learned purely from general knowledge without even requiring any unlabeled domain-specific data, these methods are called dataless classification~\cite{Chang2008ImportanceOS,Song2014OnDH,Yin2019BenchmarkingZT}.
Later, topic models~\cite{Chen2015DatalessTC,Li2016EffectiveDL} are exploited for seed-guided classification to learn seed word-aware topics by biasing the Dirichlet priors and to infer posterior document-topic assignment.
Recently, neural approaches~\cite{Mekala2020ContextualizedWS,Meng2018WeaklySupervisedNT,Meng2019WeaklySupervisedHT} have been developed for weakly-supervised text classification. They assign documents pseudo labels to train a neural classifier by either generating pseudo documents or using LMs to detect category-indicative words. While achieving inspiring performance, these neural approaches train classifiers from scratch on the local corpus and fail to take advantage of the general knowledge source used by dataless classification.
In this paper, we build our method upon pre-trained LMs, which are used both as general linguistic knowledge sources for understanding the semantics of label names, and as strong feature representation learning models for classification.

\begin{table*}[t]
\centering
\begin{tabular}{cc}
\toprule
\textbf{Sentence} & \textbf{Language Model Prediction} \\
\midrule
\makecell{The oldest annual US team \textbf{\red{sports}} competition that\\ includes professionals is not in baseball, or football or\\ basketball or hockey. It's in soccer.} & \makecell{sports, baseball, handball, soccer,\\ basketball, football, tennis, sport,\\ championship, hockey, \dots}  \\
\midrule
\makecell{Samsung's new SPH-V5400 mobile phone \textbf{\red{sports}} a built-in \\ 1-inch, 1.5-gigabyte hard disk that can store about 15 times\\ more data than conventional handsets, Samsung said.} & \makecell{has, with, features, uses, includes,\\ had, is, contains, featured, have,\\ incorporates, requires, offers, \dots}  \\
\bottomrule
\end{tabular}
\vspace{-0.5ex}
\caption{
BERT language model prediction (sorted by probability) for the word to appear at the position of ``sports'' under different contexts. The two sentences are from \textit{AG News} corpus.
}
\label{tab:word_understand}
\vspace{-0.5ex}
\end{table*}

\section{Method}

In this section, we introduce \lotclass with BERT~\cite{Devlin2019BERTPO} as our backbone model, but our method can be easily adapted to other pre-trained neural LMs.

\subsection{Category Understanding via Label Name Replacement}
\label{sec:topic}

When provided label names, humans are able to understand the semantics of each label based on general knowledge by associating with it other correlated keywords that indicate the same category. 
In this section, we introduce how to learn a category vocabulary from the label name of each class with a pre-trained LM, similar to the idea of topic mining in recent studies~\cite{Meng2020DiscriminativeTM,Meng2020HierarchicalTM}.

Intuitively, words that are interchangeable most of the time are likely to have similar meanings. We use the pre-trained BERT masked language model (MLM) to predict what words can replace the label names under most contexts. 
Specifically, for each occurrence of a label name in the corpus, we feed its contextualized embedding vector $\bs{h} \in \mathbb{R}^{h}$ produced by the BERT encoder to the MLM head, which will output a probability distribution over the entire vocabulary $V$, indicating the likelihood of each word $w$ appearing at this position:
\begin{equation}
\label{eq:lm_pred}
p(w \mid \bs{h}) = \text{Softmax}\left(W_2 \, \sigma \left( W_1 \bs{h} + \bs{b} \right) \right),
\end{equation}
where $\sigma(\cdot)$ is the activation function; $W_1 \in \mathbb{R}^{h\times h}$, $\bs{b} \in \mathbb{R}^{h}$, and $W_2 \in \mathbb{R}^{|V|\times h}$ are learnable parameters that have been pre-trained with the MLM objective of BERT.

Table~\ref{tab:word_understand} shows the pre-trained MLM prediction for the top words (sorted by $p(w \mid \bs{h})$) to replace the original label name ``sports'' under two different contexts. 
We observe that for each masked word, the top-$50$ predicted words usually have similar meanings with the original word, and thus we use the threshold of $50$ words given by the MLM to define valid replacement for each occurrence of the label names in the corpus.
Finally, we form the category vocabulary of each class using the top-$100$ words ranked by how many times they can replace the label name in the corpus, discarding stopwords with NLTK~\cite{bird2009natural} and words that appear in multiple categories. Tables~\ref{tab:keyword_vocab_agnews}, \ref{tab:keyword_vocab_imdb}, \ref{tab:keyword_vocab_amazon} and \ref{tab:keyword_vocab_dbpedia} (Table~\ref{tab:keyword_vocab_dbpedia} is in Appendix~\ref{sec:keyword_vocabs}) show the label name used for each category and the obtained category vocabulary of \textit{AG News}, \textit{IMDB}, \textit{Amazon} and \textit{DBPedia} corpora, respectively.

\begin{table*}
\centering
\begin{tabular}{cc}
\toprule
\textbf{Label Name} & \textbf{Category Vocabulary} \\
\midrule
politics & \makecell{politics, political, politicians, government, elections, politician, democracy,\\ democratic, governing, party, leadership, state, election, politically, affairs, issues,\\ governments, voters, debate, cabinet, congress, democrat, president, religion, \dots} \\
\midrule
sports & \makecell{sports, games, sporting, game, athletics, national, athletic, espn, soccer, basketball,\\ stadium, arts, racing, baseball, tv, hockey, pro, press, team, red, home, bay, kings,\\ city, legends, winning, miracle, olympic, ball, giants, players, champions, boxing, \dots} \\
\midrule
business & \makecell{business, trade, commercial, enterprise, shop, money, market, commerce, corporate,\\ global, future, sales, general, international, group, retail, management, companies,\\ operations, operation, store, corporation, venture, economic, division, firm, \dots} \\
\midrule
technology & \makecell{technology, tech, software, technological, device, equipment, hardware, devices,\\ infrastructure, system, knowledge, technique, digital, technical, concept, systems,\\ gear, techniques, functionality, process, material, facility, feature, method, \dots} \\
\bottomrule
\end{tabular}
\vspace{-0.5ex}
\caption{
The label name used for each class of \textit{AG News} dataset and the learned category vocabulary.
}
\label{tab:keyword_vocab_agnews}
\vspace{-2ex}
\end{table*}

\begin{table*}[t]
\centering
\scalebox{1.0}{
\begin{tabular}{cc}
\toprule
\textbf{Label Name} & \textbf{Category Vocabulary} \\
\midrule
good & \makecell{good, excellent, fair, wonderful, sound, high, okay, positive, sure, solid, quality,\\ smart, normal, special, successful, quick, home, brilliant, beautiful, tough, fun,\\ cool, amazing, done, interesting, superb, made, outstanding, sweet, happy, old, \dots} \\
\midrule
bad & \makecell{bad, badly, worst, mad, worse, sad, dark, awful, rotten, rough, mean, dumb,\\ negative, nasty, mixed, thing, much, fake, guy, ugly, crazy, german, gross, weird,\\ sorry, like, short, scary, way, sick, white, black, shit, average, dangerous, stuff, \dots} \\
\bottomrule
\end{tabular}
}
\caption{
The label name used for each class of \textit{IMDB} dataset and the learned category vocabulary.
}
\label{tab:keyword_vocab_imdb}
\end{table*}

\begin{table*}[t]
\centering
\scalebox{1.0}{
\begin{tabular}{cc}
\toprule
\textbf{Label Name} & \textbf{Category Vocabulary} \\
\midrule
good & \makecell{good, excellent, fine, right, fair, sound, wonderful, high, okay, sure, quality, smart,\\ positive, solid, special, home, quick, safe, beautiful, cool, valuable, normal,\\ amazing, successful, interesting, useful, tough, fun, done, sweet, rich, suitable, \dots} \\
\midrule
bad & \makecell{bad, terrible, horrible, badly, wrong, sad, worst, worse, mad, dark, awful, mean,\\ rough, rotten, much, mixed, dumb, nasty, sorry, thing, negative, funny, far, go, crazy,\\ weird, lucky, german, shit, guy, ugly, short, weak, sick, gross, dangerous, fake, \dots} \\
\bottomrule
\end{tabular}
}
\caption{
The label name used for each class of \textit{Amazon} dataset and the learned category vocabulary.
}
\label{tab:keyword_vocab_amazon}
\vspace{-1ex}
\end{table*}


\subsection{Masked Category Prediction}
\label{sec:mcp}

\begin{figure*}[t]
\centering
\includegraphics[width=1.0\textwidth]{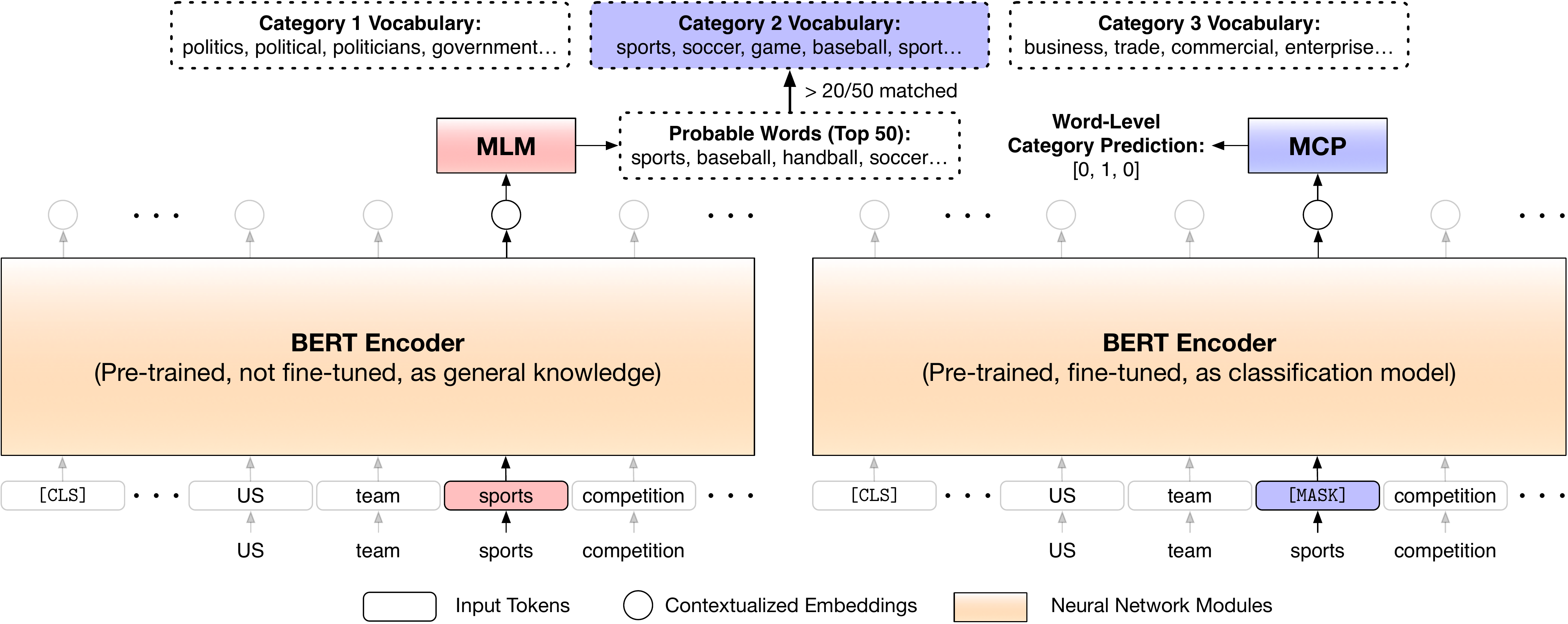}
\caption{Overview of Masked Category Prediction (MCP). The Masked Language Model (MLM) head first predicts what are probable words to appear at each token's position. A token is considered as ``category-indicative'' if its probable replacement words highly overlap with the category vocabulary of a certain class. The MCP head is trained to predict the implied categories of the category-indicative words with them masked.} 
\label{fig:model}
\end{figure*}

Like how humans perform classification, we want the classification model to focus on category-indicative words in a sequence. A straightforward way is to directly highlight every occurrence of the category vocabulary entry in the corpus. However, this approach is error-prone because: (1) Word meanings are contextualized; not every occurrence of the category keywords indicates the category. For example, as shown in Table~\ref{tab:word_understand}, the word ``sports'' in the second sentence does not imply the topic ``sports''. (2) The coverage of the category vocabulary is limited; some terms under specific contexts have similar meanings with the category keywords but are not included in the category vocabulary.

To address the aforementioned challenge, we introduce a new task, Masked Category Prediction (MCP), as illustrated in Fig.~\ref{fig:model}, wherein a pre-trained LM creates \emph{contextualized} word-level category supervision for training itself to predict the implied category of a word with the word masked.


To create contextualized word-level category supervision, we reuse the pre-trained MLM method in Section~\ref{sec:topic} to understand the contextualized meaning of each word by examining what are valid replacement words. As shown in Table~\ref{tab:word_understand}, the MLM predicted words are good indicators of the original word's meaning. As before, we regard the top-$50$ words given by the MLM as valid replacement of the original word, and we consider a word $w$ as ``category-indicative'' for class $c_w$ if more than $20$ out of $50$ $w$'s replacing words appear in the category vocabulary of class $c_w$. By examining every word in the corpus as above, we will obtain a set of category-indicative words and their category labels $\mathcal{S}_{\text{ind}}$ as word-level supervision.

For each category-indicative word $w$, we mask it out with the \texttt{[MASK]} token and train the model to predict $w$'s indicating category $c_w$ via cross-entropy loss with a classifier (a linear layer) on top of $w$'s contextualized embedding $\bs{h}$:
\begin{align}
\label{eq:mcp}
\mathcal{L}_{MCP} &= - \sum_{(w, c_w) \in \mathcal{S}_{\text{ind}}} \log p(c_w \mid \bs{h}_w), \\
\label{eq:cls}
p(c \mid \bs{h}) &= \text{Softmax}\left(W_c \bs{h} + \bs{b}_c \right),
\end{align}
where $W_c \in \mathbb{R}^{K \times h}$ and $\bs{b}_c \in \mathbb{R}^{K}$ are learnable parameters of the linear layer ($K$ is the number of classes).

We note that it is crucial to mask out the category-indicative word for category prediction, because this forces the model to infer categories based on the word's \emph{contexts} instead of simply memorizing context-free category keywords. In this way, the BERT encoder will learn to encode category-discriminative information within the sequence into the contextualized embedding $\bs{h}$ that is helpful for predicting the category at its position.


\subsection{Self-Training}

After training the LM with the MCP task, we propose to self-train the model on the entire unlabeled corpus for two reasons: (1) There are still many unlabeled documents not seen by the model in the MCP task (due to no category keywords detected) that can be used to refine the model for better generalization. (2) The classifier has been trained on top of words to predict their categories with them masked, but have not been applied on the \texttt{[CLS]} token where the model is allowed to see the entire sequence to predict its category. 

The idea of self-training (ST) is to iteratively use the model's current prediction $P$ to compute a target distribution $Q$ which guides the model for refinement. The general form of ST objective can be expressed with the KL divergence loss:
\begin{equation}
\label{eq:st}
\mathcal{L}_{ST} = \text{KL} (Q \| P) = \sum_{i=1}^{N} \sum_{j=1}^{K} q_{ij} \log \frac{q_{ij}}{p_{ij}},
\end{equation}
where $N$ is the number of instances.

There are two major choices of the target distribution $Q$: Hard labeling and soft labeling. Hard labeling~\cite{Lee2013PseudoLabelT} converts high-confidence predictions over a threshold $\tau$ to one-hot labels, \ie, $q_{ij} = \mathbbm{1}(p_{ij} > \tau)$, where $\mathbbm{1}(\cdot)$ is the indicator function. Soft labeling~\cite{Xie2016UnsupervisedDE} derives $Q$ by enhancing high-confidence predictions while demoting low-confidence ones via squaring and normalizing the current predictions:
\begin{equation}
\label{eq:soft_label}
q_{ij} = \frac{p_{ij}^2 / f_j}{\sum_{j'} \left( p_{ij'}^2 / f_{j'} \right)},\, f_j = \sum_i p_{ij},
\end{equation}
where the model prediction is made by applying the classifier trained via MCP (Eq.~\eqref{eq:cls}) to the \texttt{[CLS]} token of each document, \ie, 
\begin{equation}
\label{eq:cls_pred}
p_{ij} = p(c_j \mid \bs{h}_{d_i:\texttt{[CLS]}}).
\end{equation}

In practice, we find that the soft labeling strategy consistently gives better and more stable results than hard labeling, probably because hard labeling treats high-confident predictions directly as ground-truth labels and is more prone to error propagation. Another advantage of soft labeling is that the target distribution is computed for every instance and no confidence thresholds need to be preset.

We update the target distribution $Q$ via Eq.~\eqref{eq:soft_label} every $50$ batches and train the model via Eq.~\eqref{eq:st}. The overall algorithm is shown in Algorithm~\ref{alg:train}.

\begin{algorithm}[h]
\caption{\lotclass Training.}
\label{alg:train}
\KwIn{
An unlabeled text corpus $\mathcal{D}$; a set of label names $\mathcal{C}$; a pre-trained neural language model $M$.
}
\KwOut{A trained model $M$ for classifying the $K$ classes.}

Category vocabulary $\gets$ Section~\ref{sec:topic}\;
$\mathcal{S}_{\text{ind}} \gets$ Section~\ref{sec:mcp}\;
Train $M$ with Eq.~\eqref{eq:mcp}\;
$B \gets$ Total number of batches\;
\For{$i \gets 0$ to $B-1$}
{
\If{$i \bmod 50 = 0$} {
$Q \gets$ Eq.~\eqref{eq:soft_label}\;
}
Train $M$ on batch $i$ with Eq.~\eqref{eq:st}\;
}
Return $M$\;
\end{algorithm}


\section{Experiments}

\subsection{Datasets}

\begin{table*}[t]
\centering
\begin{tabular}{*{5}{c}}
\toprule
\textbf{Dataset} & \textbf{Classification Type} & \textbf{\# Classes} & \textbf{\# Train} & \textbf{\# Test} \\
\midrule
\textbf{AG News} & News Topic & 4 & 120,000 & 7,600 \\
\textbf{DBPedia} & Wikipedia Topic & 14 & 560,000 & 70,000 \\
\textbf{IMDB} & Movie Review Sentiment & 2 & 25,000 & 25,000 \\
\textbf{Amazon} & Product Review Sentiment & 2 & 3,600,000 & 400,000 \\
\bottomrule
\end{tabular}
\caption{
Dataset statistics. Supervised models are trained on the entire training set. Semi-supervised models use $10$ labeled documents per class from the training set and the rest as unlabeled data. Weakly-supervised models are trained by using the entire training set as unlabeled data. All models are evaluated on the test set.
}
\label{tab:dataset}
\end{table*}

We use four benchmark datasets for text classification: \textit{AG News}~\cite{Zhang2015CharacterlevelCN}, \textit{DBPedia}~\cite{Lehmann2015DBpediaA}, \textit{IMDB}~\cite{Maas2011LearningWV} and \textit{Amazon}~\cite{McAuley2013HiddenFA}. The dataset statistics are shown in Table~\ref{tab:dataset}. All datasets are in English language.

\subsection{Compared Methods}

We compare \lotclass with a wide range of weakly-supervised methods and also state-of-the-art semi-supervised and supervised methods. The label names used as supervision on each dataset for the weakly-supervised methods are shown in Tables~\ref{tab:keyword_vocab_agnews}, \ref{tab:keyword_vocab_imdb}, \ref{tab:keyword_vocab_amazon} and \ref{tab:keyword_vocab_dbpedia}. (Table~\ref{tab:keyword_vocab_dbpedia} can be found in Appendix~\ref{sec:keyword_vocabs}.) Fully supervised methods use the entire training set for model training. Semi-supervised method \textbf{UDA} uses $10$ labeled documents per class from the training set and the rest as unlabeled data. Weakly-supervised methods use the training set as unlabeled data. All methods are evaluated on the test set.

\paragraph{Weakly-supervised methods:}
\begin{itemize}[leftmargin=*]
\item \textbf{Dataless}~\cite{Chang2008ImportanceOS}: Dataless classification maps label names and each document into the same semantic space of Wikipedia concepts. Classification is performed based on vector similarity between documents and
classes using explicit semantic analysis~\cite{Gabrilovich2007ComputingSR}.

\item \textbf{WeSTClass}~\cite{Meng2018WeaklySupervisedNT}: WeSTClass generates pseudo documents to pre-train a CNN classifier and then bootstraps the model on unlabeled data with self-training.

\item \textbf{BERT w. simple match}: We treat each document containing the label name as if it is a labeled document of the corresponding class to train the BERT model.

\item \textbf{\lotclass w/o. self train}: This is an ablation version of our method. We train \lotclass only with the MCP task, without performing self-training on the entire unlabeled data.
\end{itemize}

\paragraph{Semi-supervised method:}
\begin{itemize}[leftmargin=*]
\item \textbf{UDA}~\cite{Xie2019UnsupervisedDA}: Unsupervised data augmentation is the state-of-the-art semi-supervised text classification method. Apart from using a small amount of labeled documents for supervised training, it uses back translation~\cite{Sennrich2016ImprovingNM} and TF-IDF word replacing for augmentation and enforces the model to make consistent predictions over the augmentations.
\end{itemize}

\paragraph{Supervised methods:}
\begin{itemize}[leftmargin=*]
\item \textbf{char-CNN}~\cite{Zhang2015CharacterlevelCN}: Character-level CNN was one of the state-of-the-art supervised text classification models before the appearance of neural LMs. It encodes the text sequences into characters and applies $6$-layer CNNs for feature learning and classification.

\item \textbf{BERT}~\cite{Devlin2019BERTPO}: We use the pre-trained BERT-base-uncased model and fine-tune it with the training data for classification.
\end{itemize}

\subsection{Experiment Settings}

We use the pre-trained BERT-base-uncased model as the base neural LM. For the four datasets \textit{AG News}, \textit{DBPedia}, \textit{IMDB} and \textit{Amazon}, the maximum sequence lengths are set to be $200$, $200$, $512$ and $200$ tokens. The training batch size is $128$. We use Adam~\cite{Kingma2015AdamAM} as the optimizer. The peak learning rate is $2e-5$ and $1e-6$ for MCP and self-training, respectively. The model is run on $4$ NVIDIA GeForce GTX 1080 Ti GPUs.

\subsection{Results}

\begin{table*}[t]
\centering
\begin{tabular}{ll*{5}{c}}
\toprule
\textbf{Supervision Type} & \textbf{Methods} & \textbf{AG News} & \textbf{DBPedia} & \textbf{IMDB} & \textbf{Amazon} \\
\midrule
\multirow{5}{*}{\textbf{Weakly-Sup.}}
& \textbf{Dataless}~\cite{Chang2008ImportanceOS} & 0.696 & 0.634 & 0.505 & 0.501 \\ 
& \textbf{WeSTClass}~\cite{Meng2018WeaklySupervisedNT} & 0.823 & 0.811 & 0.774 & 0.753 \\
& \textbf{BERT w. simple match} & 0.752 & 0.722 & 0.677 & 0.654 \\
& \textbf{\lotclass w/o. self train} & 0.822 & 0.860 & 0.802 & 0.853\\
& \lotclass & \textbf{0.864} & \textbf{0.911} & \textbf{0.865} & \textbf{0.916} \\
\midrule
\multirow{1}{*}{\textbf{Semi-Sup.}}
& \textbf{UDA}~\cite{Xie2019UnsupervisedDA} & 0.869 & 0.986 & 0.887 & 0.960 \\ 
\midrule
\multirow{2}{*}{\textbf{Supervised}}
& \textbf{char-CNN}~\cite{Zhang2015CharacterlevelCN} & 0.872 & 0.983 & 0.853 & 0.945 \\
& \textbf{BERT}~\cite{Devlin2019BERTPO} & 0.944 & 0.993 & 0.945 & 0.972 \\
\bottomrule
\end{tabular}
\caption{
Test accuracy of all methods on four datasets.
}
\label{tab:acc}
\end{table*}

The classification accuracy of all methods on the test set is shown in Table~\ref{tab:acc}. \lotclass consistently outperforms all weakly-supervised methods by a large margin. Even without self-training, \lotclass's ablation version performs decently across all datasets, demonstrating the effectiveness of our proposed category understanding method and the MCP task. With the help of self-training, \lotclass's performance becomes comparable to state-of-the-art semi-supervised and supervised models.

\paragraph{How many labeled documents are label names worth?} We vary the number of labeled documents per class on \textit{AG News} dataset for training \textbf{Supervised BERT}  and show its corresponding performance in Fig.~\ref{fig:worth}. The performance of \lotclass is equivalent to that of \textbf{Supervised BERT} with $48$ labeled documents per class.

\subsection{Study of Category Understanding}

\begin{table*}[hbt!]
\centering
\scalebox{1.0}{
\begin{tabular}{cc}
\toprule
\textbf{Label Name} & \textbf{Category Vocabulary} \\
\midrule
commerce & \makecell{commerce, trade, consumer, retail, trading, merchants, treasury, currency, sales,\\ commercial, market, merchant, economy, economic, marketing, store, exchange,\\ transactions, marketplace, businesses, investment, markets, trades, enterprise, \dots} \\
\midrule
economy & \makecell{economy, economic, economies, economics, currency, trade, future, gdp, treasury,\\ sector, production, market, investment, growth, mortgage, commodity, money,\\ markets, commerce, economical, prosperity, account, income, stock, store, \dots} \\
\bottomrule
\end{tabular}
}
\caption{
Different label names used for class ``business'' of \textit{AG News} dataset and the learned category vocabulary.
}
\label{tab:sensitivity}
\end{table*}
\begin{table*}[hbt!]
\centering
\scalebox{1.0}{
\begin{tabular}{cc}
\toprule
\textbf{Label Name} & \textbf{Category Vocabulary} \\
\midrule
good & \makecell{good, better, really, always, you, well, excellent, very, things, think, way, sure, \\ thing, so, n't, we, lot, get, but, going, kind, know, just, pretty, i, 'll, certainly, 're,\\ nothing, what, bad, great, best, something, because, doing, got, enough, even, \dots} \\
\midrule
bad & \makecell{bad, good, things, worse, thing, because, really, too, nothing, unfortunately, awful,\\ n't, pretty, maybe, so, lot, trouble, something, wrong, got, terrible, just, anything, \\kind, going, getting, think, get, ?, you, stuff, 've, know, everything, actually, \dots} \\
\bottomrule
\end{tabular}
}
\caption{
GloVe 300-d pre-trained embedding for category understanding on \textit{Amazon} dataset.
}
\label{tab:alter}
\end{table*}
We study the characteristics of the method introduced in Section~\ref{sec:topic} from the following two aspects. (1) Sensitivity to different words as label names. 
We use ``commerce'' and ``economy'' to replace ``business'' as the label name on \textit{AG News} dataset. Table~\ref{tab:sensitivity} shows the resulting learned category vocabulary. We observe that despite the change in label name, around half of terms in the resulting category vocabulary overlap with the original one (Table~\ref{tab:keyword_vocab_agnews} ``business'' category); the other half also indicate very similar meanings.
This guarantees the robustness of our method since it is the category vocabulary rather than the original label name that is used in subsequent steps. (2) Advantages over alternative solutions. We take the pre-trained $300$-d GloVe~\cite{Pennington2014GloveGV} embeddings and use the top words ranked by cosine similarity with the label names for category vocabulary construction. On \textit{Amazon} dataset, we use ``good'' and ``bad'' as the label names, and the category vocabulary built by \lotclass (Table~\ref{tab:keyword_vocab_amazon}) accurately reflects the sentiment polarity, while the results given by GloVe (Table~\ref{tab:alter}) are poor---some words that are close to ``good''/``bad'' in the GloVe embedding space do not indicate sentiment, or even the reversed sentiment (the closest word to ``bad'' is ``good''). This is because context-free embeddings only learn from local context windows, while neural LMs capture long-range dependency that leads to accurate interpretation of the target word.

\subsection{Effect of Self-Training}
\begin{figure*}[h]
\subfigcapmargin=10pt
\centering
\subfigure[\textbf{Supervised BERT}: Test acc. vs. number of labeled documents.]{
	\label{fig:worth}
	\includegraphics[width = 0.312\textwidth]{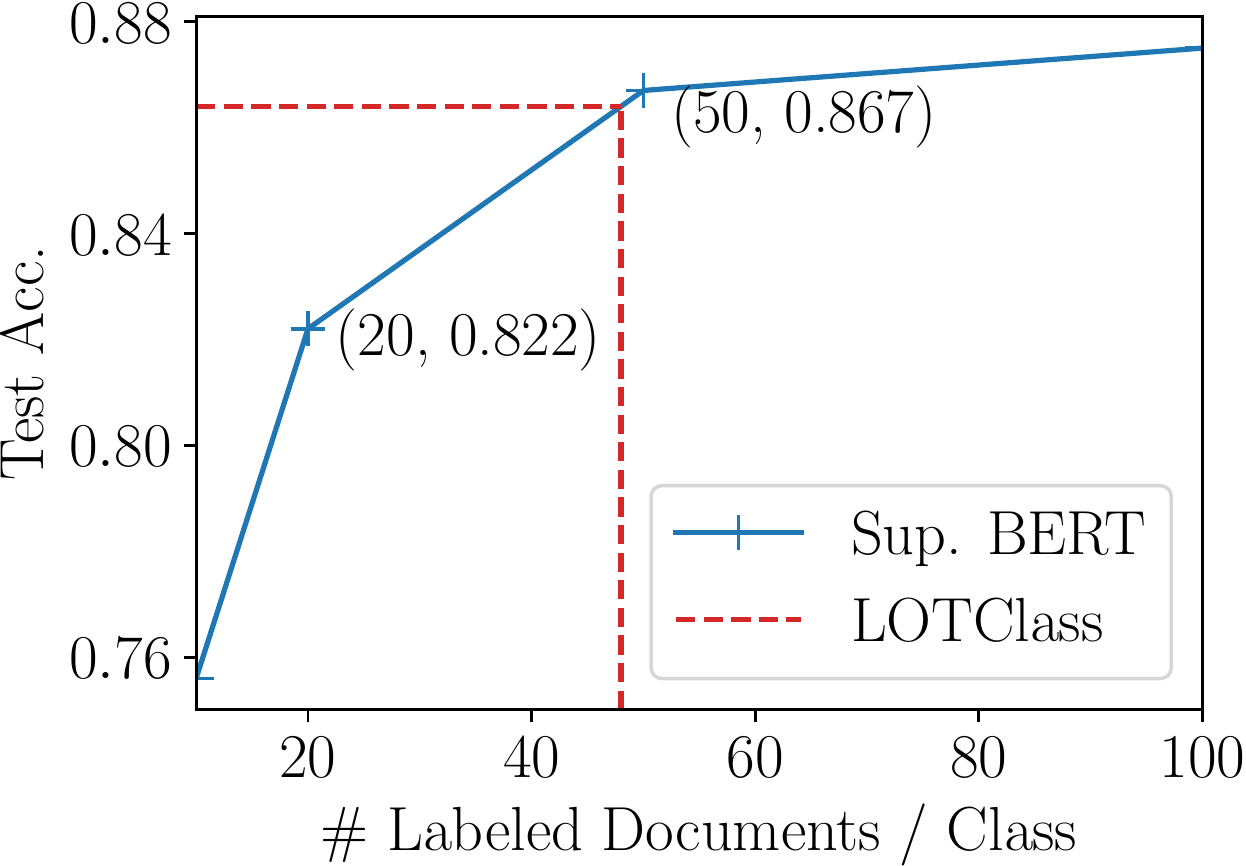}
}
\subfigure[\lotclass: Test accuracy and self-training loss.]{
	\label{fig:acc_loss}
	\includegraphics[width = 0.344\textwidth]{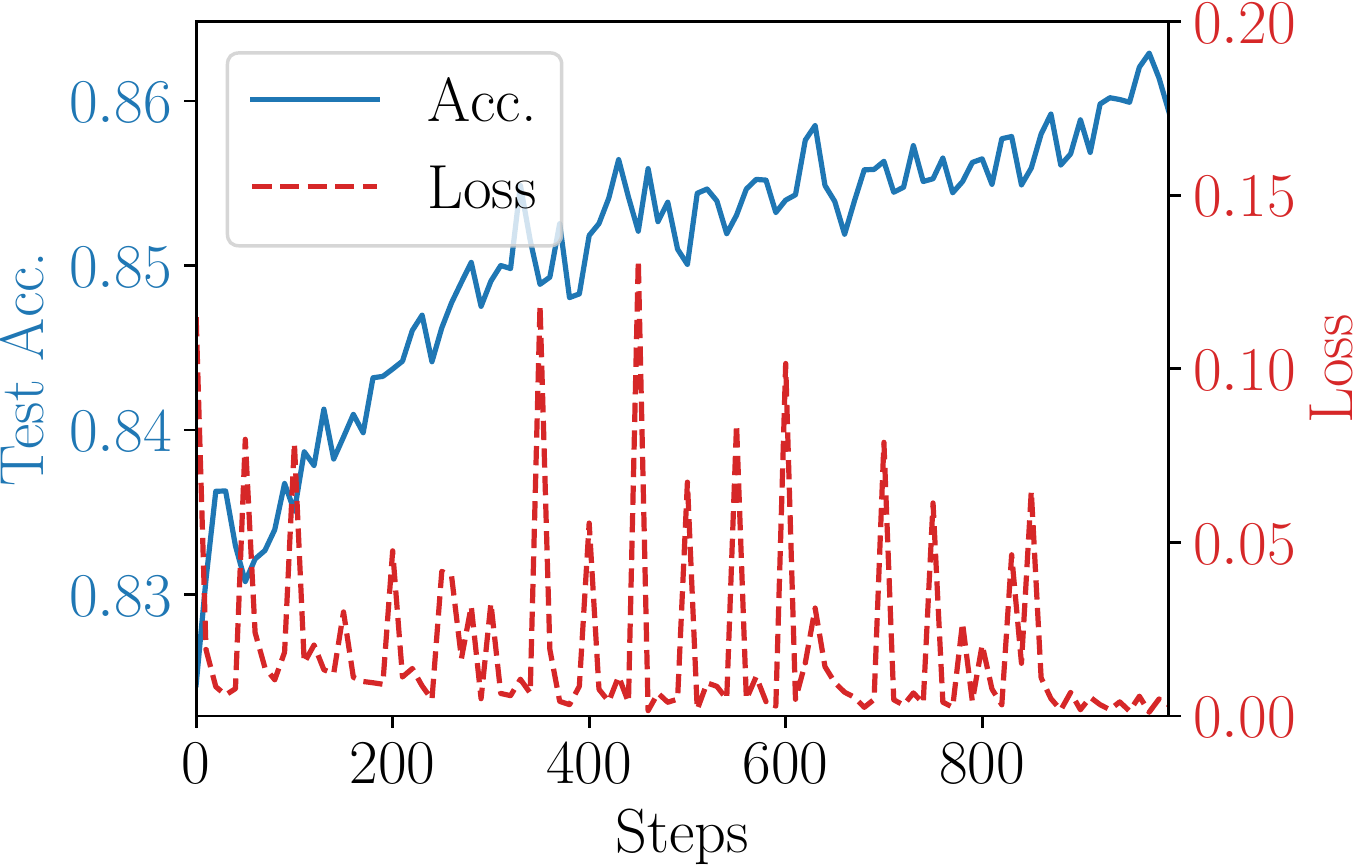}
}
\subfigure[\lotclass vs. \textbf{BERT w. simple match} during self-training.]{
	\label{fig:compare}
	\includegraphics[width = 0.295\textwidth]{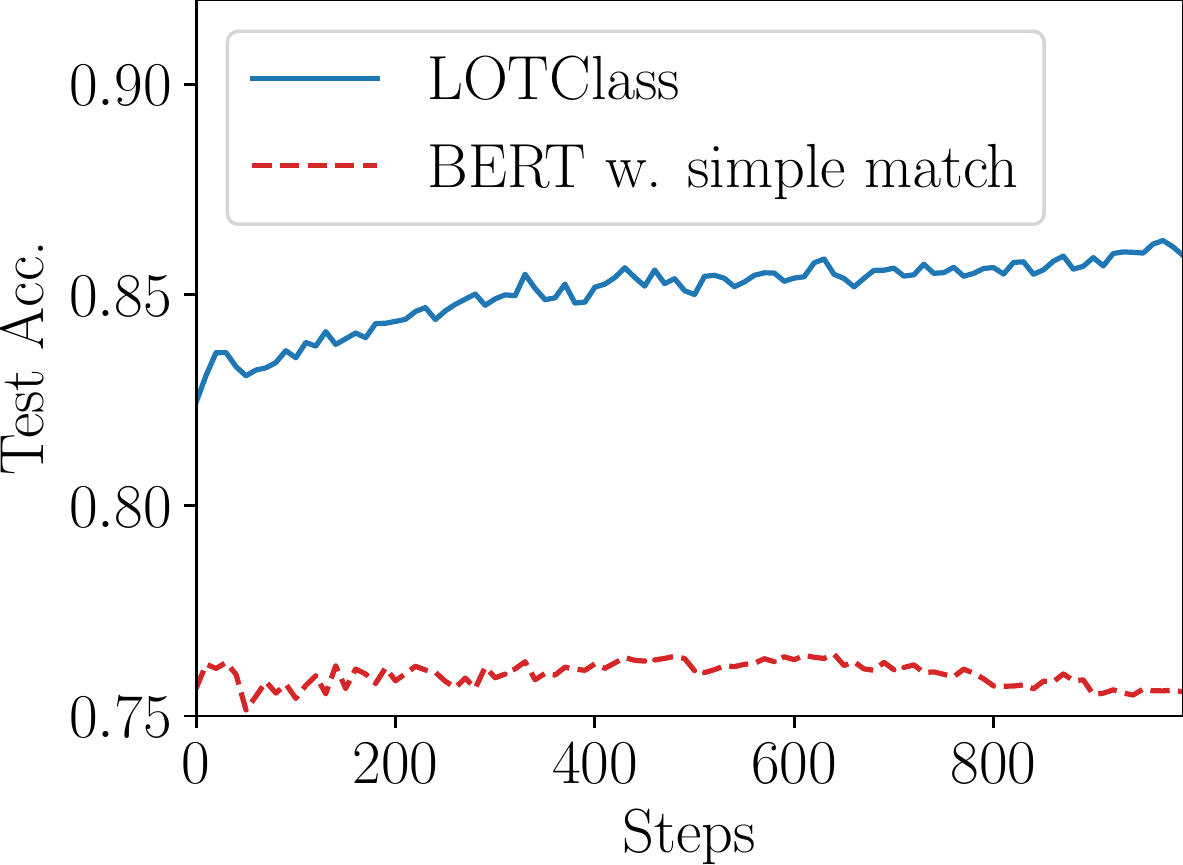}
}
\caption{(On \textit{AG News} dataset.) (a) The performance of \lotclass is close to that of \textbf{Supervised BERT} with $48$ labeled documents per class. (b) The self-training loss of \lotclass decreases in a period of $50$ steps; the performance of \lotclass gradually improves. (c) \textbf{BERT w. simple match} does not benefit from self-training.
}
\label{fig:self_train}
\end{figure*}
We study the effect of self-training with two sets of experiments: (1) In Fig.~\ref{fig:acc_loss} we show the test accuracy and self-training loss (Eq.~\eqref{eq:st}) when training \lotclass on the first $1,000$ steps (batches) of unlabeled documents. It can be observed that the loss decreases within a period of $50$ steps, which is the update interval for the target distribution $Q$---when the self training loss approximates zero, the model has fit the previous $Q$ and a new target distribution is computed based on the most recent predictions. With the model refining itself on unlabeled data iteratively, the performance gradually improves. (2) In Fig.~\ref{fig:compare} we show the performance of \lotclass vs. \textbf{BERT w. simple match} with the same self-training strategy. \textbf{BERT w. simple match} does not seem to benefit from self-training as our method does. This is probably because documents containing label names may not be actually about the category (\eg, the second sentence in Table~\ref{tab:word_understand}); the noise from simply matching the label names causes the model to make high-confidence wrong predictions, from which the model struggles to extract correct classification signals for self-improvement. This demonstrates the necessity of creating word-level supervision by understanding the contextualized word meaning and training the model via MCP to predict the category of words instead of directly assigning the word's implied category to its document.


\section{Discussions}

\paragraph{The potential of weakly-supervised classification has not been fully explored.}
For the simplicity and clarity of our method, (1) we only use the BERT-base-uncased model rather than more advanced and recent LMs; (2) we use at most $3$ words per class as label names; (3) we refrain from using other dependencies like back translation systems for augmentation. We believe that the performance will become better with the upgrade of the model, the enrichment in inputs and the usage of data augmentation techniques.

\paragraph{Applicability of weak supervision in other NLP tasks.}
Many other NLP problems can be formulated as classification tasks such as named entity recognition and aspect-based sentiment analysis~\cite{Huang2020AspectBasedSA}. Sometimes a label name could be too generic to interpret (\eg, ``person'', ``time'', etc). To apply similar methods as introduced in this paper to these scenarios, one may consider instantiating the label names with more concrete example terms like specific person names.

\paragraph{Limitation of weakly-supervised classification.}
There are difficult cases where label names are not sufficient to teach the model for correct classification. For example, some review texts implicitly express sentiment polarity that goes beyond word-level understanding:
\textit{``I find it sad that just because Edward Norton did not want to be in the film or have anything to do with it, people automatically think the movie sucks without even watching it or giving it a chance.''}
Therefore, it will be interesting to improve weakly-supervised classification with active learning where the model is allowed to consult the user about difficult cases.

\paragraph{Collaboration with semi-supervised classification.} 
One can easily integrate weakly-supervised methods with semi-supervised methods in different scenarios: (1) When no training documents are available, the high-confidence predictions of weakly-supervised methods can be used as ground-truth labels for initializing semi-supervised methods. (2) When both training documents and label names are available, a joint objective can be designed to train the model with both word-level tasks (\eg, MCP) and document-level tasks (\eg, augmentation, self-training).


\section{Conclusions}

In this paper, we propose the \lotclass model built upon pre-trained neural LMs for text classification with label names as the only supervision in three steps: Category understanding via label name replacement, word-level classification via masked category prediction, and self-training on unlabeled corpus for generalization. The effectiveness of \lotclass is validated on four benchmark datasets. 
We show that label names is an effective supervision type for text classification but has been largely overlooked by the mainstreams of literature. 
We also point out several directions for future work by generalizing our methods to other tasks or combining with other techniques.

\section*{Acknowledgments}
Research was sponsored in part by US DARPA KAIROS Program No.\ FA8750-19-2-1004 and SocialSim Program No.\ W911NF-17-C-0099, National Science Foundation IIS 19-56151, IIS 17-41317, IIS 17-04532, IIS 16-18481, and III-2008334, and DTRA HDTRA11810026. Any opinions, findings, and conclusions or recommendations expressed herein are those of the authors and should not be interpreted as necessarily representing the views, either expressed or implied, of DARPA or the U.S. Government. The U.S. Government is authorized to reproduce and distribute reprints for government purposes notwithstanding any copyright annotation hereon. 
We thank anonymous reviewers for valuable and insightful feedback.
\bibliographystyle{acl_natbib}
\bibliography{ref}


\appendix

\section{Label Names Used and Category Vocabulary Obtained for DBPedia}
\label{sec:keyword_vocabs}

We show the label names used for \textit{DBPedia} corpora and the obtained category vocabulary in Table~\ref{tab:keyword_vocab_dbpedia}. In most cases, only one word as the label name will be sufficient; however, sometimes the semantics of the label name might be too general so we instead use $2$ or $3$ keywords of the class to represent the label name. For example, we use ``school'' and ``university'' to represent the class ``educational institution''; we use ``river'', ``lake'' and ``mountain'' to represent the class ``natural place''; we use ``book'', ``novel'' and ``publication'' to represent the class ``paper work''.

\onecolumn
\begin{longtable}{cc}
\toprule
\textbf{Label Name} & \textbf{Category Vocabulary} \\
\midrule
company & \makecell{companies, co, firm, concern, subsidiary, brand, enterprise, division, partnership,\\ manufacturer, works, inc, cooperative, provider, corp, factory, chain, limited,\\ holding, consortium, industry, manufacturing, entity, operator, product, giant \dots} \\
\midrule
\makecell{school\\ university} & \makecell{academy, college, schools, ecole, institution, campus, university, secondary,\\ form, students, schooling, standard, class, educate, elementary, hs, level,\\ student, tech, academic, universities, branch, degree, universite, universidad, \dots} \\
\midrule
artist & \makecell{artists, painter, artistic, musician, singer, arts, poet, designer, sculptor, composer,\\ star, vocalist, illustrator, architect, songwriter, entertainer, cm, painting,\\ cartoonist, creator, talent, style, identity, creative, duo, editor, personality, \dots} \\
\midrule
athlete & \makecell{athletes, athletics, indoor, olympian, archer, events, sprinter, medalist, olympic,\\ runner, jumper, swimmer, competitor, holder, mile, ultra, able, mark, hurdles,\\ relay, amateur, medallist, footballer, anchor, metres, cyclist, shooter, athletic, \dots} \\
\midrule
politics & \makecell{politics, political, government, politicians, politician, elections, policy, party,\\ affairs, legislature, politically, democracy, democratic, governing, history,\\ leadership, cabinet, issues, strategy, election, religion, assembly, law, \dots} \\
\midrule
transportation & \makecell{transportation, transport, transit, rail, travel, traffic, mobility, bus, energy,\\ railroad, communication, route, transfer, passenger, transported, traction,\\ recreation, metro, shipping, railway, security, transports, infrastructure, \dots}\\
\midrule
building & \makecell{buildings, structure, tower, built, wing, hotel, build, structures, room,\\ courthouse, skyscraper, library, venue, warehouse, block, auditorium, location,\\ plaza, addition, museum, pavilion, landmark, offices, foundation, headquarters, \dots}\\
\midrule
\makecell{river\\ lake\\ mountain} & \makecell{river, lake, bay, dam, rivers, water, creek, channel, sea, pool, mountain,\\ stream, lakes, flow, reservoir, hill, flowing, mountains, basin, great, glacier,\\ flowed, pond, de, valley, peak, drainage, mount, summit, brook, mare, head, \dots}\\
\midrule
village & \makecell{village, villages, settlement, town, east, population, rural, municipality, parish,\\ na, temple, commune, pa, ha, north, pre, hamlet, chamber, settlements, camp,\\ administrative, lies, township, neighbourhood, se, os, iran, villagers, nest, \dots}\\
\midrule
animal & \makecell{animal, animals, ape, horse, dog, cat, livestock, wildlife, nature, lion, human,\\ owl, cattle, cow, wild, indian, environment, pig, elephant, fauna, mammal,\\ beast, creature, australian, ox, land, alligator, eagle, endangered, mammals, \dots}\\
\midrule
\makecell{plant\\ tree} & \makecell{shrub, plants, native, rose, grass, herb, species, jasmine, race, vine, hybrid,\\ bamboo, hair, planted, fire, growing, flame, lotus, sage, iris, perennial, variety,\\ palm, cactus, trees, robert, weed, nonsense, given, another, stand, holly, poppy, \dots}\\
\midrule
\makecell{album} & \makecell{lp, albums, cd, ep, effort, recording, disc, compilation, debut, appearance,\\ soundtrack, output, genus, installation, recorded, anthology, earth, issue, imprint, ex,\\ era, opera, estate, single, outing, arc, instrumental, audio, el, song, offering, \dots}\\
\midrule
\makecell{film} & \makecell{films, comedy, drama, directed, documentary, video, language, pictures,\\ miniseries, negative, movies, musical, screen, trailer, acting, starring, filmmaker,\\ flick, horror, silent, screenplay, box, lead, filmmaking, second, bond, script, \dots}\\
\midrule
\makecell{book\\ novel\\ publication} & \makecell{novel, books, novels, mystery, memoir, fantasy, fiction, novelist, reader, read, cycle,\\ romance, writing, written, published, novella, play, narrative, trilogy, manga,\\ autobiography, publication, literature, isbn, write, tale, poem, year, text, reading, \dots}\\
\bottomrule
\caption{
The label name used for each class of \textit{DBPedia} dataset and the learned category vocabulary.
}
\label{tab:keyword_vocab_dbpedia}
\end{longtable}

\end{document}